%% file: FAGEN.tex
\documentclass{article}
\usepackage{microtype}
\usepackage{graphicx}
\usepackage{booktabs}
\usepackage{xurl}
\usepackage{hyperref}
\usepackage{tcolorbox}
\usepackage[accepted]{icml2026}

\usepackage{amsmath}
\usepackage{amssymb}
\icmltitlerunning{Instruction Bleed: Cross-Module Interference in Prompt-Composed Agentic Systems}

\begin{document}

\twocolumn[
  \icmltitle{Instruction Bleed: Cross-Module Interference \\
    in Prompt-Composed Agentic Systems}

  \begin{icmlauthorlist}
    \icmlauthor{Ching-Yu Lin}{uiuc}
    \icmlauthor{Yifan Liu}{uiuc}
  \end{icmlauthorlist}

  \icmlaffiliation{uiuc}{University of Illinois Urbana-Champaign, Urbana, IL, USA}

  \icmlcorrespondingauthor{Yifan Liu}{yifan40@illinois.edu}

  \icmlkeywords{agentic systems, failure modes, cross-module interference, agent evaluation, reliability}

  \vskip 0.3in
]

\printAffiliationsAndNotice{}

\begin{abstract}
Practitioners of prompt-composed agentic systems report a recurring failure mode: editing one prompt module silently shifts the behavior of others despite no shared variable or executable dependency.
We formalize this as \emph{compositional behavioral leakage} (CBL): interference between modules sharing a context window. 
CBL is enabled by architectural non-isolation: transformer self-attention provides no formal boundary between concatenated modules.
We probe CBL on a deployed job-evaluation agent (Claude Sonnet~4.6, 144 trials) through a reusable three-channel protocol that perturbs non-focal modules along \emph{volume}, \emph{content}, and \emph{form}.
Only the \emph{content} channel produces a detectable paired effect (Cohen's $d = 0.63$, bootstrap 95\% CI excluding zero); no recommendation flipped---a sub-threshold regime invisible to standard QA but compounding across the thousands of decisions a deployed agent makes. 
CBL is orthogonal to known agent-failure axes (adversarial injection, cognitive degradation, multi-agent fault propagation, privacy leakage). 
We contribute an operational definition, a reusable protocol, a falsifiable prediction set, and a system-class characterization, establishing cross-module interference measurement as a requirement for prompt-composed agent evaluation.
\end{abstract}

\section{Introduction}
\label{sec:intro}

Prompt-composed agentic systems---where an agent's behavior is assembled
at runtime from natural-language modules rather than written as
executable code, and a large language model (LLM) interprets the
composed context as the agent's policy---are a fast-growing deployment
pattern. Practitioners of these systems already report a recurring
failure mode: editing one prompt module silently shifts the behavior of
an unrelated module, with no shared variable, tool call, or executable
dependency to explain it~\citep{subrahmanyam2025gemini}. The locus of
risk is therefore the composition mechanism itself, not any single
module. OpenClaw~\yrcite{openclaw2026} typifies this deployment pattern
at production scale: 379K GitHub stars and over 10K community-authored
skill modules, with users composing personality definitions (SOUL.md),
capability modules (SKILL.md), and orchestration rules (AGENTS.md) into
agent configurations that no single author fully audits.

The same pattern recurs across mechanism
choices---markdown convention~\citep{openclaw2026,careerops2026},
template engines~\citep{openhands2024}, and Python class
inheritance~\citep{aider2023}---with non-trivial behavioral logic
carried in text the LLM interprets and code restricted to deterministic
I/O (\S\ref{sec:systems}, Table~\ref{tab:systems}).

The architectural assumption that composing prompt modules yields the
composition of their individual behaviors is not warranted by
transformer self-attention, which computes global pairwise interactions
across the composed context with no module-level isolation: a
\texttt{\#\#\# Persona} heading is a writing convention, not a namespace
boundary. Proactive interference~\citep{wang2025unable} and
coverage-bounded compositional
generalization~\citep{chang2025pattern} jointly predict that each novel
module combination is a distribution shift whose behavioral consequences
are not derivable from per-module evaluation.

Yet no existing agent benchmark measures whether modifying one prompt
module silently alters a semantically unrelated module's
behavior---making compositional behavior under module-set shift a
first-order safety and reliability concern for deployed agents.

We make four contributions:
\begin{itemize}
  \item We give an operational definition of \emph{compositional behavioral leakage}
    (CBL)---behavioral interference between co-resident prompt modules in a
    shared attention context---and distinguish it from compositional
    privacy leakage~\citep{patil2025sum}, longitudinal cognitive
    degradation, and adversarial prompt injection.
  \item We instantiate a reusable three-channel protocol on a deployed
    job-evaluation agent (Claude Sonnet~4.6; 12~JDs, 144~trials), providing
    an existence proof of cross-module interference ($d=0.63$ on the
    semantic-content channel; condition~C2, Table~\ref{tab:results})---running
    on artifacts other prompt-composed systems already produce.
  \item We derive a falsifiable prediction set framing the multi-system
    replication program, requiring no new measurement infrastructure beyond
    what current agent deployments emit.
  \item We place prompt-composed agentic systems on a
    \emph{composition-mechanism spectrum} across four open-source projects,
    establishing them as a distinct system class within agentic systems.
\end{itemize}

\section{Prompt-Composed Agentic Systems}
\label{sec:systems}

\subsection{The Pattern}

A \textbf{prompt-composed agentic system} is one in which non-trivial
behavioral decision logic---scoring rubrics, workflow sequencing, persona
definitions, tool-selection criteria---is encoded in human-authored text
files an LLM \emph{interprets} via attention rather than executes via a
runtime.

\subsection{Four Systems}

Table~\ref{tab:systems} surveys four open-source prompt-composed agent
systems spanning a composition-mechanism spectrum from markdown
convention~\citep{openclaw2026,careerops2026} through Jinja2
templating~\citep{openhands2024} to Python class
inheritance~\citep{aider2023}. In every case behavioral logic resides in
text the LLM interprets via attention; code handles only deterministic
I/O. As systems mature, composition migrates outward (conventions $\to$
templates $\to$ class inheritance $\to$ prompt-management
services)---tacit acknowledgment that text-level composition is fragile.
Yet the most widely deployed systems (OpenClaw, AGENTS.md/CLAUDE.md
across thousands of repositories) still compose at the text level.
Appendix~\ref{app:systems-extended} provides per-system descriptions
and the full maturation pattern. The absence of a formal characterization
of when and how this fragility manifests motivates the definition we
introduce in \S\ref{sec:cbl}.

\begin{table*}[t]
\caption{Four prompt-composed agentic systems spanning a composition-mechanism spectrum. None asserts behavioral output of prompts in CI.}
\label{tab:systems}
\begin{center}
\begin{small}
\begin{tabular}{lllll}
\toprule
\textbf{System} & \textbf{Format} & \textbf{Composition} & \textbf{Files} & \textbf{Beh.\ tests} \\
\midrule
OpenClaw   & Markdown & Convention                 & SOUL/SKILL + 10K skills       & Manual only    \\
career-ops & Markdown & Convention                 & 18 modes + \texttt{CLAUDE.md}  & Struct.\ lints \\
OpenHands  & Jinja2   & \texttt{\{\% include \%\}} & 51 \texttt{.j2}                & Rendering only \\
aider      & Python   & Inheritance                & 19 \texttt{\_prompts.py}       & Parser tests   \\
\bottomrule
\end{tabular}
\end{small}
\end{center}
\vskip -0.1in
\end{table*}

\section{The Failure Mode: Compositional Behavioral Leakage}
\label{sec:cbl}

\subsection{Operational Definition}

Let $\mathcal{M}=\{m_1,\ldots,m_n\}$ be a deployed prompt-module set
concatenated into an LLM's input context, and let
$\mathcal{M}^{(j \leftarrow e)}$ denote the same module set after applying
an edit $e$ to a non-focal module $m_j$. Let $s_i(x;\mathcal{M})$ denote the
observable score or decision associated with a focal module $m_i$ on input
$x$.

We say the system exhibits \textbf{compositional behavioral leakage} (CBL)
when an edit $e$ that is not intended to affect the focal behavior induces a
detectable paired shift:
\[
\Delta_i(e)
=
\mathbb{E}_{x}
\left[
s_i(x;\mathcal{M}^{(j \leftarrow e)})
-
s_i(x;\mathcal{M})
\right],
\]
where the paired-effect bootstrap confidence interval excludes zero.

The stricter singleton comparison
$\mathcal{B}(m_i \mid \mathcal{M}) \approx \mathcal{B}(m_i \mid \{m_i\})$
is a compositional-consistency test and can be viewed as a special case of
this regression view. In this paper, we focus on the deployed-system setting:
whether localized edits to non-focal prompt modules silently shift focal
module behavior under the full deployed prompt.

CBL differs from three adjacent failure modes.
\textbf{Compositional privacy leakage}~\citep{patil2025sum}: multi-agent
output aggregation reveals sensitive information (privacy vs.\ behavior;
multi-agent vs.\ single-agent topology).
\textbf{Cognitive degradation}~\citep{atta2025qsaf}: longitudinal state drift
across multi-step execution (runtime accumulation vs.\ initialization
interference).
\textbf{Prompt injection}: adversarial external inputs subvert behavior
(malicious vs.\ accidental). Practitioners refer to CBL informally as
``instruction bleed''~\citep{subrahmanyam2025gemini}; we formalize it.

\subsection{Mechanisms Predicting CBL}

Four properties of transformer LLMs make CBL plausible and yield testable predictions.
Each contributes a distinct interference mechanism; together they make composition
\emph{syntax} itself a behavioral variable.

\paragraph{Attention is global.}
Transformer self-attention computes pairwise token interactions across the
entire input. There is no architectural mechanism restricting attention to
tokens within a single prompt module: a \texttt{\#\#\# Persona} heading is a
statistical hint, not a scope boundary, and its effectiveness depends
entirely on training-distribution priors over similar formatting in
pre-training data. Delimiter-based isolation is therefore a behavioral
expectation, not a guarantee. This non-isolation property does not by itself prove leakage in any particular
system, but it removes the formal basis for assuming that delimiter-separated
modules behave independently.

\paragraph{Proactive interference.}
Wang and Sun~\yrcite{wang2025unable} demonstrate that LLM retrieval accuracy
declines log-linearly with accumulated interfering context, even when the
target is positioned immediately before the query. Prompt-engineering
mitigations (``ignore earlier input,'' explicit scoping instructions) yield
limited effectiveness. Earlier prompt modules \emph{actively degrade} the
processing of later modules---a working-memory bottleneck that single-module
testing cannot detect, since the interference manifests only under
composition.

\paragraph{Coverage-bounded composition.}
Chang et al.~\yrcite{chang2025pattern} formalize that pattern-matching
success in compositional tasks correlates with training-distribution
coverage (cf.\ Dziri et al.~\yrcite{dziri2023faith}). Novel module
combinations are unlikely to have coverage in the pre-training distribution,
forcing fallback to priors that may blend unrelated modules. For
safety-critical deployments, this implies that untested module combinations
must be treated as behaviorally unpredictable until isolation mechanisms are
validated.

\paragraph{Format sensitivity.}
Sclar et al.~\yrcite{sclar2023quantifying} show that meaning-preserving
format changes produce accuracy swings of up to 76 percentage points within
a single model. In prompt-composed agent systems, the \emph{syntax} of
module composition is itself a behavioral variable: heading hierarchies,
list formatting, whitespace, and section ordering are interference channels
independent of semantic content. Condition~C3 in \S\ref{sec:cbl} probes this
channel directly.

\subsection{Existence Proof on a Deployed Agent}

We test for CBL in career-ops, a prompt-composed job-evaluation agentic system. The focal
module scores job descriptions on a 1--5 rubric across multiple weighted
dimensions. We measure \emph{cv\_match}, the dimension most directly
capturing the module's core evaluation function: how well a candidate's
background aligns with the job requirements.

\paragraph{Conditions.} Against an unmodified baseline (C0: the deployed system
prompt as shipped), we define three conditions that each perturb
the non-focal modules along a distinct channel:
\textbf{C1} (module addition --- \emph{volume}): baseline $+$ an unrelated
200-line recipe-evaluation module.
\textbf{C2} (semantic modification --- \emph{content}): baseline with a
semantically irrelevant archetype appended to the shared rules file
(Appendix~\ref{app:prompt-example}).
\textbf{C3} (format perturbation --- \emph{form}): baseline with
meaning-preserving structural changes (heading levels, emoji markers, section
reordering) to non-focal modules.

\textbf{C2 is the primary test:} the edit is localized to a non-focal behavioral specification, while the measured outcome is the focal cv-match score. Under the system's stated modular semantics, the added archetype should not systematically change this score; a paired shift therefore indicates cross-module interference.

\paragraph{Primary outcome.}
We pre-specify CV-match as the primary outcome on construct-validity
grounds: 
it is the rubric dimension most directly downstream of the
Archetype Detection table modified in C2, so cross-module interference,
if present, has its highest theoretical prior of manifesting on this
dimension.

\paragraph{Protocol.} Each condition is evaluated on 12 job descriptions
$\times$ 3 independent runs using Claude Sonnet~4.6 (144 total trials).
Effect sizes (Cohen's $d$) with bootstrap 95\% CIs on $\Delta$
(10{,}000 resamples) serve as the primary analysis; Wilcoxon signed-rank
tests with Bonferroni correction provide secondary confirmation.

\paragraph{Results.} Table~\ref{tab:results} reports per-condition \emph{cv\_match}
scores and paired effect sizes with bootstrap 95\% CIs.

\begin{table}[t]
\caption{Per-condition cv-match scores relative to C0 (baseline; mean\,=\,2.72, SD\,=\,1.23). \textbf{C2} (semantic interference, primary test) shows a bootstrap 95\% CI excluding zero; \textbf{C1} and \textbf{C3} (controls) do not. Effect sizes (Cohen's $d$) computed on JD-level paired means ($N\!=\!12$); bootstrap 95\% CIs on $\Delta$ from 10{,}000 resamples.}
\label{tab:results}
\begin{center}
\begin{small}
\setlength{\tabcolsep}{4pt}
\begin{tabular}{@{}lrrrrr@{}}
\toprule
\textbf{Cond.} & \textbf{Mean} & \textbf{SD} & $\boldsymbol{\Delta}$ & $\boldsymbol{d}$ & \textbf{95\% CI ($\Delta$)} \\
\midrule
C1 (volume)    & 2.69 & 1.19 & $-0.03$ & $-0.11$       & $[-0.19,\,+0.11]$  \\
C2 (content)   & 2.89 & 1.17 & $+0.17$ & \textbf{0.63} & $[+0.03,\,+0.31]$  \\
C3 (form)      & 2.64 & 1.27 & $-0.08$ & $-0.29$       & $[-0.25,\,+0.08]$  \\
\bottomrule
\end{tabular}
\end{small}
\end{center}
\vskip -0.1in
\end{table}

\textbf{C2 is the strongest signal:} adding a single irrelevant archetype to
the shared rules file produces $d = 0.63$ ($\Delta = {+}0.17$, bootstrap 95\%
CI $[+0.03,\,+0.31]$, excluding zero), with 8 of 12 JDs shifting upward.
Within-condition ICC for cv-match is 0.925, so paired contrasts at $N = 12$
retain power for the medium-to-large effect sizes the framework predicts under
semantic interventions. 
This shift satisfies our operational criterion for CBL: a localized edit to a
non-focal behavioral specification induces a detectable shift in the focal
cv-match score.

\textbf{The control conditions sharpen the claim.} \textbf{C1}
($d = -0.11$) and \textbf{C3} ($d = -0.29$) yield CIs including zero,
ruling out both a generic ``any-added-context degrades performance''
interpretation and a format-channel explanation, and localizing the
detected interference to semantic content---sharpening rather than
weakening the CBL claim. No recommendation flips occurred under any
condition---a sub-threshold regime the framework predicts under small
semantic interventions, developed in \S\ref{sec:implications}.

\section{Related Work}
\label{sec:related}

\paragraph{Agent failure-mode evaluation.}
AgentDojo~\yrcite{debenedetti2024agentdojo} evaluates adversarial injection;
QSAF~\yrcite{atta2025qsaf} proposes runtime mitigation for cognitive
degradation; MAS-FIRE~\yrcite{masfire2026} evaluates multi-agent fault
propagation; Patil et al.~\yrcite{patil2025sum} evaluate compositional
\emph{privacy} leakage. None measure synchronous behavioral interference
between co-resident prompt modules in a single agent's initialization
context.

\paragraph{Prompt sensitivity and context interference.}
Sclar et al.~\yrcite{sclar2023quantifying} document accuracy swings from
meaning-preserving format changes;
PI-LLM~\yrcite{wang2025unable} shows log-linear retrieval decline resistant to
prompt-engineering mitigations. Our contribution extends these from
controlled tasks into real agent behavioral output.

\paragraph{Modular prompting.}
DecomP~\yrcite{khot2022decomposed} decomposes tasks via multi-call
orchestration; ModularPrompt~\yrcite{chen2022learning},
PRopS~\yrcite{pilault2023conditional}, and Smart MoP~\yrcite{dun2023smart}
achieve modularity through soft prompts and differentiable or softmax gating.
Each isolates modules through explicit mathematical or API-level mechanisms;
the systems we study compose at the text level via concatenation, with no
isolation beyond attention.

\paragraph{Compositional generalization.}
Skill-Mix~\yrcite{yu2024skillmix} shows frontier models combine skills but
degrade as skill count increases. Chang et al.~\yrcite{chang2025pattern}
formalize that compositional generalization is bounded by
training-distribution coverage; Dziri et al.~\yrcite{dziri2023faith} show
transformers collapse into pattern matching under compositional depth.
Model-level compositional capacity is therefore insufficient to guarantee
clean composition when novel module combinations lack coverage.

\section{Deployed-System Risks and Predictions}
\label{sec:implications}

\paragraph{Deployed-system risks.}
CBL's practical consequence is quiet unreliability, not loud failure: in our
experiment no recommendation flipped while scores shifted systematically.
Under continuous prompt evolution---community-contributed skills,
accumulating CLAUDE.md rules, user-customized archetypes---such silent drift
is hard to detect and harder to audit.

\paragraph{Sub-threshold drift as predicted regime.}
C2's result lies in the regime CBL predicts under small semantic
interventions: score-based outputs drift before decision boundaries cross.
Sub-threshold drift is invisible to standard QA (which checks decisions, not
score distributions), compounds across the thousands of decisions a deployed
agent makes, and propagates into downstream ranking, prioritization, and
aggregation systems where magnitude matters independently of individual
flips. Detecting it \emph{requires} protocols like the C0--C3
framework---the methodological contribution of this paper.

\paragraph{Falsifiable predictions.}
The CBL framework yields three falsifiable predictions for the multi-system
replication program.
\emph{(i)~Cross-model variation.} Magnitude of cross-module interference
should vary substantially across model families, given that within-model
format perturbations alone produce large accuracy (up to 76 percentage points)
swings~\citep{sclar2023quantifying}; this makes model-migration testing a
first-class requirement.
\emph{(ii)~Semantic-distance gradient.} Shifts on rubric dimensions
semantically proximate to the modified module should systematically exceed
shifts on distal dimensions, providing a graded interpretability test of
the attentional-pathway hypothesis.
\emph{(iii)~Threshold-crossing scaling.} Recommendation-flip rates should
scale with intervention magnitude and with cross-module semantic overlap.
All three are falsifiable on artifacts other prompt-composed agentic
systems already produce, with no new measurement infrastructure required.

\paragraph{Scope and open questions.}
The case study uses one model and one agentic system---appropriate for a
proof-of-concept demonstration of the protocol; the falsifiable predictions
above frame the multi-model, multi-system replication program. Whether model
providers can offer module-isolation primitives---e.g., separately
cached prompt segments with restricted cross-segment attention---or whether
global attention makes text-level isolation fundamentally impossible remains
open. CBL likely extends to multimodal agents and web-browsing agents that
compose instructions with crawled content.

\subsection{Regression Testing Framework}
\label{sec:regression}

Prompt modules treated as software components require regression testing
across four categories: \emph{compositional consistency},
\emph{module-interaction regression}, \emph{format-perturbation robustness},
and \emph{model-migration testing}~\citep{chang2025pattern,sclar2023quantifying}.
Conditions C1--C3 instantiate the latter three, providing a reusable
protocol; none of the systems in \S\ref{sec:systems} implements any of
these tests. Practitioners' informal ``prompt sprawl''~\citep{subrahmanyam2025gemini}
acquires an operational definition.

\paragraph{Compositional consistency.}
Behavior of a focal module under composition with the full module set
should approximate its behavior under singleton context:
$\mathcal{B}(m_i \mid \mathcal{M}) \approx \mathcal{B}(m_i \mid \{m_i\})$
across representative module sets. Deviations indicate cross-module
interference and should be quantified by paired effect sizes with
bootstrap confidence intervals, as instantiated by the C0-vs-C2 contrast
in \S\ref{sec:cbl}.

\paragraph{Module-interaction regression.}
When a module is added to an existing module set, the behavioral suite
for all existing modules must be re-evaluated. Condition~C1 in
\S\ref{sec:cbl} instantiates this test: the focal module's behavior is
measured under (existing modules) versus (existing $+$ added module),
detecting whether the addition silently shifts non-focal-module
behavior.

\paragraph{Format-perturbation robustness.}
Module behavior should survive meaning-preserving structural changes
(heading levels, list formatting, whitespace, section ordering) to
co-resident modules. Condition~C3 in \S\ref{sec:cbl} instantiates this
test. Format perturbation is a documented first-class behavioral
variable, not a nuisance
variable~\citep{sclar2023quantifying}.

\paragraph{Model-migration testing.}
Coverage-bounded composition~\citep{chang2025pattern} makes interference
magnitude training-distribution-dependent, so behavioral drift must be
re-measured under model updates. Cross-model differences exceed
within-model format perturbations~\citep{sclar2023quantifying},
predicting at least comparable variation. Behavioral suites pinned to a
specific model version must be re-run on every model upgrade rather
than assumed transferable.

\paragraph{Scaling considerations.}
Pairwise interaction testing scales as $O(n^2)$---prohibitive for
OpenClaw's 10K+ community-authored skills, which yield over 50
million pairwise combinations before considering higher-order
interactions. Practical testing strategies must therefore use
sampling-based approaches: random pair sampling for monitoring, semantic-
or attention-distance-weighted sampling for high-risk pairs, or
covering-design approaches drawn from combinatorial software testing.
None of the systems in \S\ref{sec:systems} currently implements any
form of regression test along any of these categories.

\section{Conclusion}
\label{sec:conclusion}

Compositional behavioral leakage is a measurable failure mode in
prompt-composed agentic systems; the protocol we introduce provides an
existence proof on a deployed agent and a falsifiable prediction set for
broader replication. 
The phenomenon is consistent with a basic architectural non-isolation
property: absent explicit isolation mechanisms, transformer self-attention
computes interactions across the full context rather than respecting
delimiter-defined prompt-module boundaries.
In our case study, a single irrelevant archetype shifted cv-match scores by
$d = 0.63$ ($\Delta = {+}0.17$, bootstrap 95\% CI
$[+0.03,\,+0.31]$; Claude Sonnet~4.6) despite no semantic relationship to
the focal task. We deliver four artifacts---an operational definition of
CBL, a reusable three-channel protocol, a falsifiable prediction set, and a
system-class characterization of prompt-composed agentic
systems---establishing cross-module interference measurement as a
requirement for prompt-composed agent evaluation. With combinatorial
module-composition spaces no test suite can exhaustively cover (OpenClaw's
10K+ skills make untested combinations the norm), the framework's
predictions are testable on artifacts these systems already produce, with
no new measurement infrastructure required.

\section*{Impact Statement}

This paper identifies a failure mode in prompt-composed agentic systems that
may affect the reliability of score-based decisions. In applications where
such scores influence outcomes for people---hiring, credit, triage---even
sub-threshold behavioral shifts raise fairness concerns. Our contribution is
diagnostic: we formalize and measure an existing phenomenon to enable
mitigation, rather than introducing new capabilities that could be misused.

\bibliography{references}
\bibliographystyle{icml2026}

\newpage
\appendix
\onecolumn
\input{appendix_FMAI}
\end{document}

%% file: appendix_FMAI.tex
\section{Extended System Descriptions}
\label{app:systems-extended}

This appendix elaborates the four prompt-composed agent systems summarized
in Table~\ref{tab:systems}, providing the per-system descriptions and the
maturation pattern referenced from \S\ref{sec:systems}.

\paragraph{OpenClaw.}
OpenClaw~\yrcite{openclaw2026} (379K GitHub stars) exemplifies the paradigm
at maximum scale. Users compose personality files (SOUL.md), capability
modules (SKILL.md), and orchestration rules (AGENTS.md) that are
concatenated into the system prompt at startup~\citep{openclaw2026}. Over
10K community-authored skills are installable from a public registry,
with no sandboxing between skill contexts---composition is resolved
entirely by the LLM's attention mechanism.

\paragraph{career-ops.}
career-ops~\yrcite{careerops2026}, a job-evaluation pipeline, uses 18
markdown mode files with explicit cross-references. All behavioral
logic---archetype taxonomy, scoring rubric, decision thresholds---resides
in markdown; code handles only I/O.

\paragraph{OpenHands.}
OpenHands~\yrcite{openhands2024} uses Jinja2 templating to enable modular
composition, yet rendered prompts are still concatenated into a single
context window before reaching the LLM.

\paragraph{aider.}
aider~\yrcite{aider2023} encodes prompts as Python string constants on
classes that inherit from a \texttt{CoderPrompts} base. This is the most
code-like mechanism---but the strings are still natural-language
behavioral specifications interpreted via attention.

\paragraph{Maturation pattern.}
As systems mature, composition migrates out of prompts into code or
external services (conventions $\to$ templates $\to$ class inheritance
$\to$ prompt-management services)---tacit acknowledgment that text-level
composition is fragile. Yet the most widely deployed systems (OpenClaw,
AGENTS.md/CLAUDE.md across thousands of repositories) still compose at
the text level, with composition resolved at runtime by the LLM's
attention mechanism rather than by any explicit isolation primitive.

\section{Prompt Manipulation: C0 vs.\ C2}
\label{app:prompt-example}

The shared rules file consumed by the job-evaluation agent contains an
\emph{Archetype Detection} table that maps role types to keyword signals.
Condition~C0 (baseline) uses the original table; Condition~C2 appends a
semantically irrelevant ``Professional Chef'' archetype together with a
culinary-specific evaluation note.  The excerpts below show the
\texttt{Archetype Detection} section only; all other sections of the
rules file are identical across conditions.

\medskip
\noindent\textbf{C0 --- Baseline}

\begin{tcolorbox}[colback=white, colframe=black!50,
                  title={C0 --- Baseline rules file (excerpt)},
                  fonttitle=\bfseries]
{\footnotesize
\begin{verbatim}

## Archetype Detection

Classify every offer into one of these types (or hybrid of 2):

| Archetype                | Key signals in JD                               |
|--------------------------|-------------------------------------------------|
| AI Platform / LLMOps     | "observability", "evals", "pipelines", ...      |
| Agentic / Automation     | "agent", "HITL", "orchestration", ...           |
| Technical AI PM          | "PRD", "roadmap", "discovery", ...              |
| AI Solutions Architect   | "architecture", "enterprise", "integration", ...|
| AI Forward Deployed      | "client-facing", "deploy", "prototype", ...     |
| AI Transformation        | "change management", "adoption", ...            |

After detecting archetype, read _profile.md for the user's specific
framing and proof points for that archetype.
\end{verbatim}
}
\end{tcolorbox}

\medskip
\noindent\textbf{C2 --- Irrelevant-Archetype Addition}

Lines marked \texttt{[+]} are the only additions relative to C0.

\begin{tcolorbox}[colback=white, colframe=black!50,
                  title={C2 --- Modified rules file (excerpt)},
                  fonttitle=\bfseries]

\footnotesize
\begin{verbatim}
## Archetype Detection

Classify every offer into one of these types (or hybrid of 2):

| Archetype                | Key signals in JD                               |
|--------------------------|-------------------------------------------------|
| AI Platform / LLMOps     | "observability", "evals", "pipelines", ...      |
| Agentic / Automation     | "agent", "HITL", "orchestration", ...           |
| Technical AI PM          | "PRD", "roadmap", "discovery", ...              |
| AI Solutions Architect   | "architecture", "enterprise", "integration", ...|
| AI Forward Deployed      | "client-facing", "deploy", "prototype", ...     |
| AI Transformation        | "change management", "adoption", ...            |
| Professional Chef /      | "kitchen management", "menu design",          [+]
|   Culinary Specialist    |  "food safety", "culinary arts", ...          [+]

After detecting archetype, read _profile.md for the user's specific
framing and proof points for that archetype.
\end{verbatim}}
\end{tcolorbox}

\noindent The added archetype is semantically unrelated to any of the
twelve job descriptions used in the experiment, yet its presence in the
shared context shifts cv-match scores (\S\ref{sec:cbl},
Table~\ref{tab:results}).